\documentclass[letterpaper, 10 pt, conference]{IEEEtran}
\IEEEoverridecommandlockouts
\usepackage{cite}
\usepackage{amsmath,amssymb,amsfonts}
\usepackage{algorithmic}
\usepackage{graphicx}
\usepackage{textcomp}
\usepackage{xcolor}
\usepackage{multirow} 
\usepackage{booktabs} 

\usepackage[most]{tcolorbox}
\usepackage{enumitem} 
\definecolor{back}{RGB}{241, 241, 242}
\definecolor{frame}{RGB}{161, 214, 226}
\definecolor{titlebar}{RGB}{161, 214, 226}
\newtcolorbox{llmpromptbox}{
  colback=back,            
  colframe=frame,         
  coltitle=white,              
  colbacktitle=titlebar,       
  title=\textbf{LLM Prompt structure},  
  fonttitle=\normalfont\selectfont\large, 
  boxrule=0.9pt,               
  arc=4pt,                     
  left=6pt, right=6pt, top=6pt, bottom=6pt, 
  enhanced,
  sharp corners=south,
  breakable                    
}

\def\BibTeX{{\rm B\kern-.05em{\sc i\kern-.025em b}\kern-.08em
    T\kern-.1667em\lower.7ex\hbox{E}\kern-.125emX}}
\begin{document}

\title{Spatial-Semantic Reasoning using Large Language Models for Efficient UAV Search Operations
\\

\author{Marin Maletić, Marijana Peti, Tamara Petrović and Stjepan Bogdan
\thanks{Authors are with the University of Zagreb Faculty of Electrical Engineering  and Computing, LARICS (Laboratory for Robotics and Intelligent Control Systems), Unska 3, 10000 Zagreb, Croatia; {\tt\small (marin.maletic, marijana.peti, tamara.petrovic, stjepan.bogdan)@fer.hr}}
}}

\maketitle

\begin{abstract}
We present a real-time semantic navigation framework for Unmanned Aerial Vehicles (UAVs) focused on improving time efficiency in the Object Goal Navigation (ObjectNav) task. Central to our approach is a Large Language Model (LLM) that interprets user-provided natural language instructions and performs semantic reasoning over detected objects and spatial context to prioritize high-probability search regions. The system combines real-time object detection, 3D spatial mapping, and polynomial spline interpolation for smooth and feasible UAV trajectory planning. Unlike prior methods that rely on offline reasoning or simulator-constrained action spaces, our framework can operate fully onboard in real time, continuously updating semantic relevance based on new observations. Experiments in both simulated and real-world settings demonstrate reductions in mission duration while maintaining high search accuracy, underscoring the effectiveness of LLM-guided reasoning for time-efficient UAV-based ObjectNav.
\end{abstract}


\section{Introduction}
\label{sec:Introduction}

Semantic understanding of the environment and contextual reasoning are core human cognitive abilities - areas where robotic agents continue to struggle. 
Embodied Artificial Intelligence aims to bridge this gap by developing intelligent agents capable of perceiving, reasoning about, and interacting with physical or simulated environments to perform complex tasks. 
These tasks often require the integration of navigation, perception, and semantic understanding, with embodied navigation standing out as one of the most fundamental and challenging capabilities.
An important aspect of such advanced systems is also the ability to receive natural language instructions from humans.
\begin{figure}[t!]
    \centering
    \includegraphics[width=\linewidth]{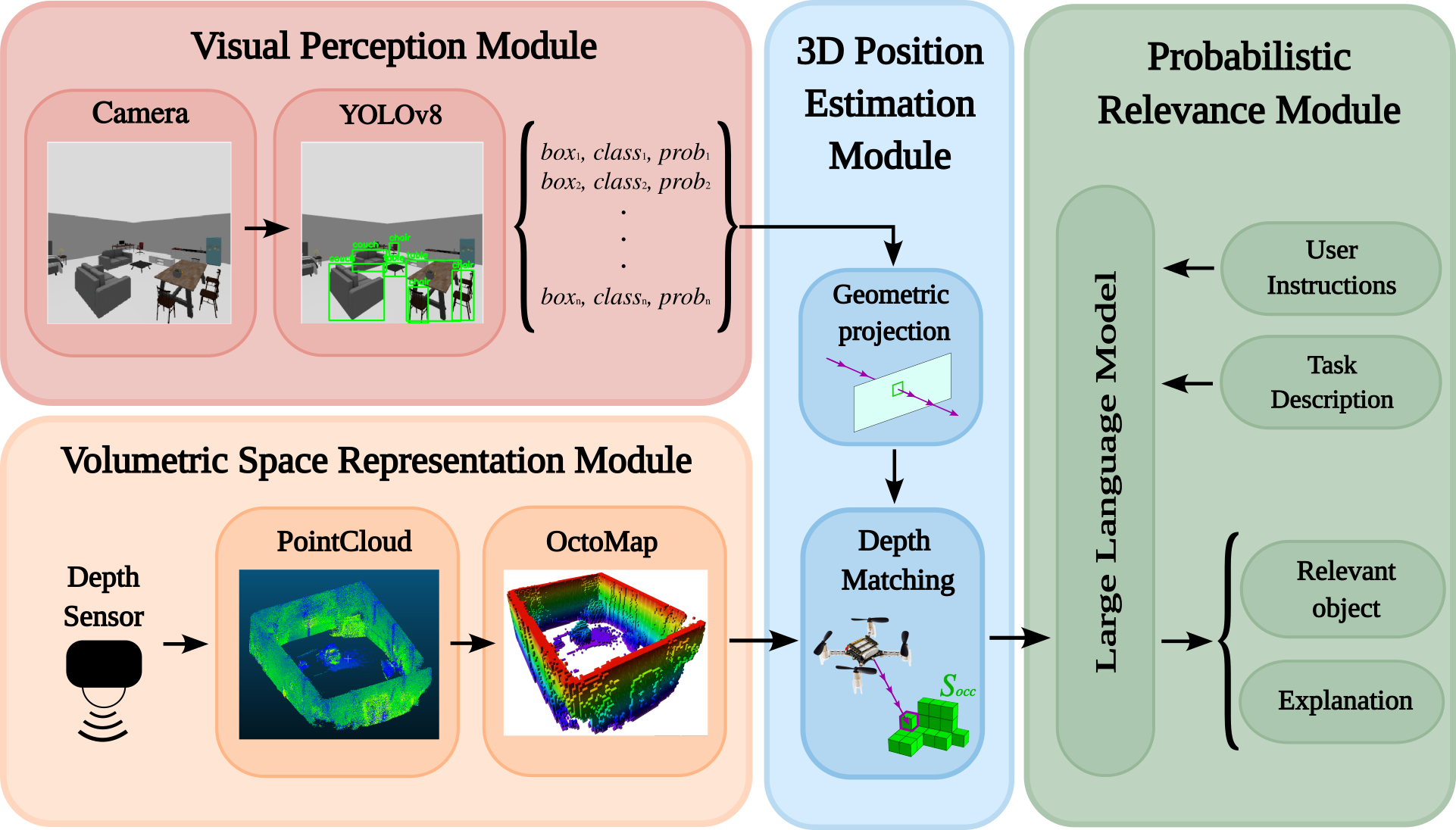}
    \caption{ Illustration of the proposed method consisting of four interconnected modules: (1) Visual Perception, employing a camera and object detection model to identify semantic objects; (2) Volumetric Space Representation, which utilizes a depth sensor to build a spatial occupancy map;  (3) Position Estimation, projecting detected objects into 3D space and associating them with depth information from the sensor data; and (4) Relevance Evaluation, which integrates all acquired semantic and spatial data with user instructions to determine the most relevant object around which the UAV performs a thorough search.}
    \label{fig:method}
\end{figure}

One such advanced task is Vision-and-Language Navigation (VLN), where agents need to interpret and execute step-by-step natural language instructions (e.g., “Turn left after the painting”) \cite{anderson2018}. 
Similarly, Object Goal Navigation (ObjectNav), which is considered in this paper, requires agents to autonomously search and locate objects specified by semantic categories (e.g., “chair”) and optionally additional description (e.g., "chair that is used in the bedroom"), within previously unseen environments \cite{objectnaw_review}. ObjectNav  was formally introduced as a core task in embodied AI by Anderson et al. \cite{anderson2018objgoal}. 


Early approaches to solving ObjectNav \cite{NEURIPS2020} typically relied on end-to-end training pipelines that required large-scale annotated datasets and substantial computational resources, limiting their scalability and generalization to real-world settings. Only recently have zero-shot methods been developed, which enable navigating to unseen targets in unknown environments.
This advancement was in large part driven by the recent progress in Large Language Models (LLMs) and Vision-Language Models (VLMs), which has significantly advanced embodied navigation by enabling robust semantic inference and natural language-based interaction with the robot. LLM-based methods, in particular, support zero-shot reasoning, eliminating the need for task-specific training, and enhancing generalization to novel scenarios. 
Recent approaches such as Zero-Shot Object Navigation (ZSON) \cite{majumdar2023zson} and its language-driven extension, Language-driven ZSON (L-ZSON) \cite{gadre2022lzson}, have demonstrated the potential of large language models (LLMs) to interpret open-ended user instructions without requiring prior exposure to specific object classes or environments. 
ZSON enables an agent to navigate toward instances of a specified object category in previously unseen environments, even in the absence of labeled training examples for that object. 
L-ZSON builds upon this foundation by incorporating natural language instructions to guide the agent’s exploration and object-seeking behavior. 
Rather than solely specifying an object category, L-ZSON tasks provide rich contextual information, such as descriptive attributes, spatial cues, or complex directives (e.g., “Find my keys near the coffee mug”), which the agent must interpret and ground in its sensory observations to effectively complete the task.

Many recent works, including OpenFMNav\cite{kuang2024openfmnav}, LM-Nav \cite{shah2022lmnav}, ESC\cite{zhou2023esc}, and PixNav\cite{cai2023pixnav}, have successfully leveraged VLMs to bridge the gap between visual perception and semantic reasoning, enabling more effective navigation in complex environments.
Methods that rely on VLMs have better performance, but the drawback is a higher computational load, often not suitable for on-board real-time inference and time-sensitive applications. 
Moreover, existing works, mainly used for ground vehicles within simulators such as Matterport3D \cite{matterport3d}, Gibson \cite{gibson}, and Habitat \cite{habitat}, use simplified discretized action spaces (e.g. forward motion, fixed-angle turns, stop). In other words, they do not take into account realistic physics-based robot maneuvers, which limits task complexity and hinders transferability to real-world systems. Only a few recent papers, such as \cite{liu2023aerialvln}, work with Unmanned Aerial Vehicles (UAVs) and take into account their realistic continuous trajectories, which is the focus of this paper. Further, existing approaches tend to prioritize spatial efficiency metrics, such as Success Rate (SR) and Success weighted by Path Length (SPL), while often overlooking practical factors like real-time responsiveness, search duration, and dynamic maneuverability. 


To address these limitations, we propose a novel framework for autonomous object goal navigation for UAVs that enables UAVs to dynamically interpret user instructions and prioritize high-likelihood search regions. 
By integrating real-time object detection with only LLM-based reasoning, we eliminate reliance on slower, compute-intensive VLMs, resulting in faster inference and greater responsiveness. 
This lightweight and modular design, illustrated in Fig.\ref{fig:method}, maintains semantic richness while enabling real-time operation. Furthermore, we use polynomial spline interpolation to generate smooth, dynamically feasible trajectories for UAVs through 3D space, allowing the UAV to continuously scan the environment and refine its reasoning during flight. Experimental results with simulated and real-world UAVs demonstrate improvements in search time and accuracy, especially for small or partially occluded objects, highlighting the practical utility and robustness of the system in real-world applications. The contribution of the paper is a novel and lightweight approach to object-goal navigation, designed for deployment on real UAVs in a zero-shot setting. The approach is language-driven and uses large language models (LLMs) to guide the search and interpret the environmental context.

The paper is divided as follows. In Section \ref{sec:Methodology} the methodology is described, including different aspects of the approach, starting from the problem formulation. Section \ref{sec:results} presents the results obtained through simulations and experiments carried out with Crazyflie UAVs \cite{crazyflie21}. The paper concludes with a summary of the findings and final remarks.

\section{Methodology}
\label{sec:Methodology}

In this paper, we introduce a novel object-search framework that leverages large language models (LLMs) to improve decision-making and efficiency in autonomous UAV search tasks. 
By employing LLMs as semantic reasoning engines, the system dynamically interprets natural language instructions and real-time object detection data, enabling intelligent path planning toward high-probability target zones, which significantly improves mission efficiency in complex environments. 
We address this challenge through a four-stage approach:
Representation of the 3D environment, Object detection and semantic grounding, LLM-guided decision logic, and Trajectory planning.

\subsection{Problem formulation}
We consider the problem of autonomously searching for a user-specified object in an unknown environment using a UAV. 
Given a natural language instruction $U$, the task is to locate the target object $T$, defined by the semantic category referenced within $U$. The unknown spatial position of the target object, $(x_T,y_T,z_T) \in \mathbb{R}^3$, is determined by the autonomous search process carried out by the UAV.
The instruction $U$ is provided as textual input containing the target object's semantic label and optionally additional contextual descriptions to guide and facilitate the search process.


The objective is to design a search method that will guide the UAV to regions where the target object is more likely to be located, thus decreasing the overall time needed to find the target object and the path taken by the agent during the search. The UAV is equipped with a camera, used for the detection of objects in the environment. The system should use high-level semantic reasoning to understand the relations between detected objects and their spatial distribution. By leveraging user specifications, contextual cues, and object co-occurrence patterns, the method should estimate the likelihood that the target object $T$ appears in the vicinity of other detected objects and prioritize search efforts accordingly. 

\subsection{Volumetric representation of the environment}
To enable safe UAV navigation and support accurate 3D localization of detected objects, a depth-aware representation of the environment is required.
\subsubsection{Volumetric Space Representation Module}
As the underlying volumetric representation, we utilize an OctoMap, a hierarchical volumetric 3D occupancy grid that provides deep understanding of the environment. 
Each cubic element of the OctoMap, called a voxel $\mathbf{v}$, can be classified as free, occupied, or unknown. 
The set of free voxels constitutes the free space $S_{free} \subset S$ in which the UAV can navigate, occupied voxels define the occupied space $S_{occ} \subset S$ that represents obstacles and objects, and unknown voxels represent the unknown space $S_{unk} \subset S$. The entire space $S$ is thus defined as the union: $S = S_{free} \cup S_{occ} \cup S_{unk} $. 
 While OctoMap is used in this work, the proposed framework is compatible with other 3D volumetric representations. 
 
 Since the focus of this paper is on LLM supported decision-making algorithms, it is assumed that the environment is mapped beforehand and $S$ is known in advance. 
 However, the proposed method can also be applied without this assumption by using sensors such as LiDAR on board the UAV, together with incremental mapping algorithms while exploring the unknown environment in real time. 
 Further, we assume that the UAV can  localize itself precisely within the mapped environment, starting from an arbitrary initial pose. 

\subsection{Object detection and semantic grounding}

As mentioned, the UAV is equipped with a camera that is used for the detection and semantic grounding of objects in the environment. In this subsection, a method for estimating the 3D positions of objects in the environment is described, which further serves as input to the LLM-guided search process. This consists of two parts - the visual perception module for detecting objects in the camera image and the 3D object position estimation module, which uses both volumetric information and the results of the visual perception module. Both modules are described as follows.

\subsubsection{Visual Perception Module}
Image frame $I_k$ is processed in real-time using a 2D visual perceptor:
\begin{equation}
   VP(I_k) = o_k,
\end{equation}
which incorporates a YOLO-based object detector. 
The visual perception module produces a set of detected objects $o_k= \{o_1, o_2, \dots, o_{m_k}\}$ for each image frame $I_k$. 
Each detected object $o_i$, for $i=1,\dots,m_k$, is represented as $\{box_i, class_i, prob_i\}$, where $box_i = (u, v, w, h)_i$ denotes the 4D bounding box vector. Here, $(u,v)$ specifies the center pixel coordinates, and $(w,h)$ indicates the width and height of the bounding box. 
Additionally, each object is assigned a semantic label $class_i$ and a detection confidence score $prob_i \in [0,1]$.

\subsubsection{3D Position Estimation Module}
\begin{figure}[]
    \centering
    \includegraphics[width=\linewidth]{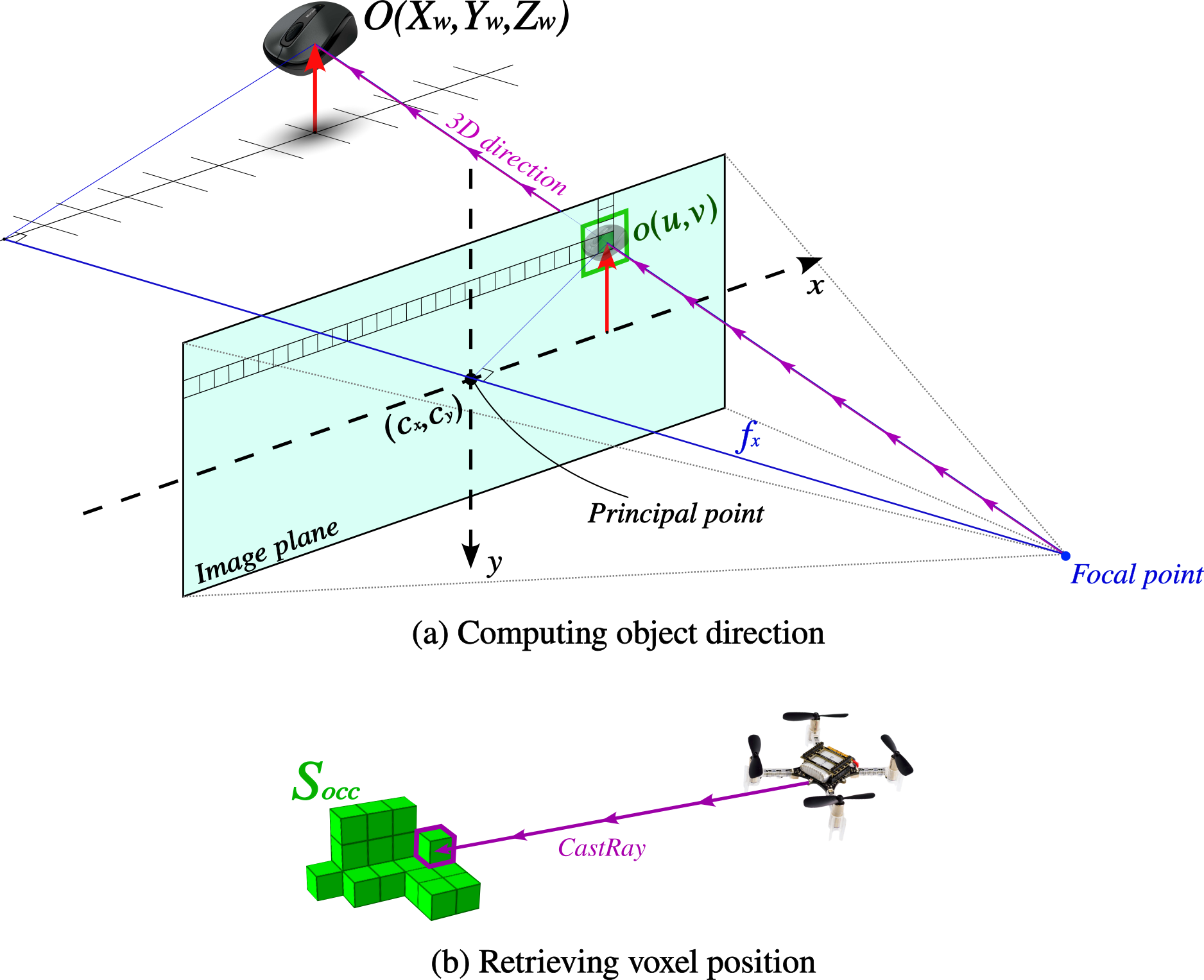}
    \caption{Illustration of the 3D position estimator $PE(\cdot)$, which maps detected objects from an image to their corresponding world coordinates in two stages: (a) Computation of the 3D viewing direction from the object’s pixel position ($u,v)$ in the image, based on camera intrinsics; (b) Ray casting from the UAV along the computed direction to identify the first occupied voxel in the OctoMap, belonging to the object.}
    \label{fig:position estimator}
\end{figure}

Leveraging the depth information provided by the occupied space $S_{occ}$ of the OctoMap, we can estimate the positions of objects detected in frame $I_k$ using the 3D position estimator: 
\begin{equation}
PE(o_k, S_{occ}) = \mathcal{O}_k,
\label{eq:position estimator}
\end{equation}
The position estimator, illustrated in Fig. \ref{fig:position estimator}, firstly utilizes the intrinsic parameters of the camera — such as resolution, lens distortion, and field of view (FOV) — to compute a 3D direction vector from the camera to each detected object based on its pixel coordinates in the image, as seen in subfigure (a). 
\begin{figure*}[t]
    \centering
    \includegraphics[width=\textwidth]{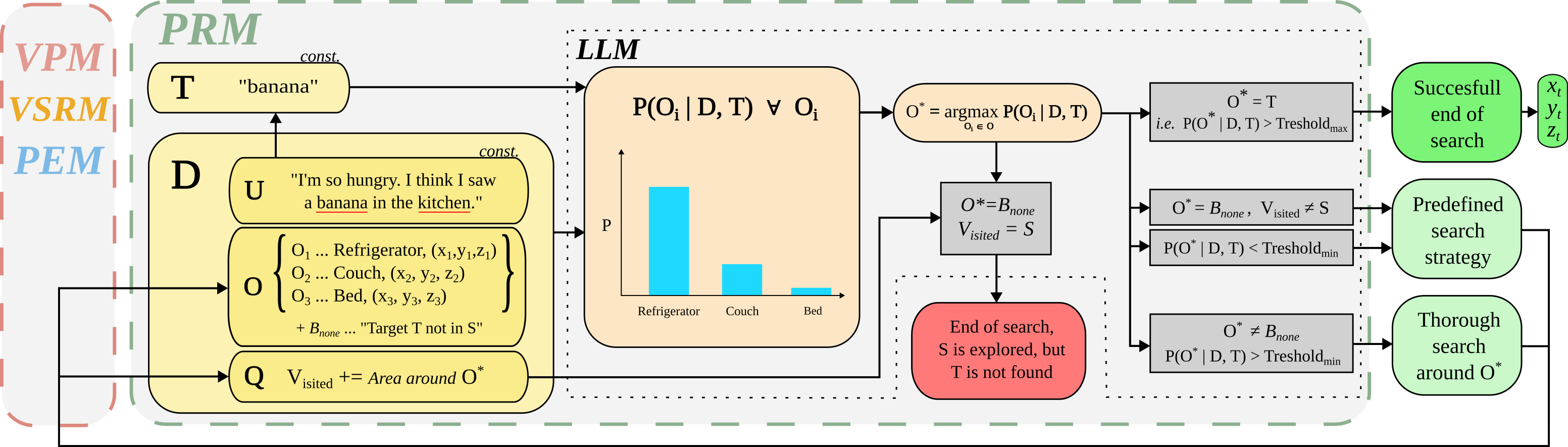}
    \caption{Overall schematic diagram of the developed system, illustrating data collection through the Visual Perception Module (VPM), Volumetric Space Representation Module (VSRM) and 3D Position Estimation Module (PEM), with emphasis on the Probabilistic Relevance Module (PRM) and the internal LLM black box. The Large Language Model determines the next search action based on accumulated data $D$, including user instructions $U$, previously searched areas $V_{isited}$, and the current set of detected objects $\mathcal{O}$ along with their spatial relationships. The LLM selects the most relevant object $O^*$; if its relevance exceeds a defined threshold, a thorough search is conducted around it. Otherwise, the system reverts to a fallback strategy (e.g., wall-following). If the relevance score for an object is sufficiently high, the LLM concludes that $O^*$ is the target $T$ and terminates the search successfully. If the UAV exhaustively searches the space $S$ without locating $T$, the $B_{none}$ hypothesis becomes dominant, and the search ends in failure.}
    \label{fig:overal_shema}
\end{figure*}
Specifically, for a detected object with center pixel $(u,v)$, $PE(\cdot)$ calculates a direction vector $d=(x_d,y_d,z_d)$ representing a ray originating from the camera’s optical center and passing through the object's center in the image, while accounting for lens distortion. 
This direction vector is then used to perform a ray-casting operation from the UAVs current world pose $C$ into the OctoMap, retrieving the first occupied voxel along the ray, which is assumed to belong to the detected object.

Consequently, we obtain $\mathcal{O}_k= \{O_1, O_2, \dots, O_{m_k}\}$, a set of $m_k$ detected objects, where each object $O_i$ is represented as $ \{pos_i, class_i\}$ for $i=1,\dots,m_k$ with $pos_i \in S_{occ}$ denoting the 3D position of the voxel within the occupied space, and $class_i$ the corresponding semantic label.

The system performs object detection and 3D position estimation for each incoming frame $I_k$ and accumulates the data.
As the UAV explores the environment and observes objects from different viewpoints, the same object may be detected multiple times from varying angles, resulting in $\sum_{k=1}^{K} m_k$ spatially scattered position estimates, particularly for larger objects, where $K$ is the number of frames. 
To address this, we apply the DBSCAN\cite{ester1996dbscan} clustering algorithm, grouping detections based on spatial proximity and shared semantic labels:
\begin{equation}
    \mathcal{O}= Cluster(\bigcup_{k=1}^{K}\mathcal{O}_k)
\end{equation}
This process consolidates multiple observations into unified object representations, yielding a final set $\mathcal{O}$ consisting of $n\ll \sum_{k=1}^{K} m_k$ distinct objects with consistent 3D positions and semantic labels.

\subsection{LLM-guided decision logic}
To enable high-level semantic and spatial reasoning in the object search process, we employ a large language model (LLM) to interpret user instructions and evaluate the relevance of detected objects with respect to the target object. \\
\subsubsection{Probabilistic Relevance Module}
We define a probabilistic relevance: 
\begin{equation}
PR(\mathcal{O},T, D) \rightarrow O^*,
\end{equation}
where $\mathcal{O}$ is a set of detected objects with their labels and spatial positions, $T$ is the target object defined by the user instructions, and $D$ is the data UAV obtained during the search, including user instructions, detected objects and their spatial relations, and visited areas.
The goal is to produce a probability distribution over $n$ detected objects $\{O_1,O_2,\dots, O_n\}$ indicating which object offers the best place to begin searching around in order to find the target object $T$ and then select that object as the most relevant, denoted as $O^*$. 
Here, we use the term \textit{relevance} to denote how likely we are to find the target object $T$ in the vicinity of a given object $O_i$. 


In other words, for each detected object $O_i$, we aim to compute:
\begin{equation} 
 P(O_i \mid D, T) \propto P(D \mid O_i, T) \cdot P(O_i \mid T)
\end{equation}
where $ P(O_i \mid D, T)$ represents the \textit{relevancy score}, \textit{i.e.} posterior probability that object $O_i$ is relevant given the target $T$ and the current knowledge $D$.
Here, based on Bayes rule, $P(O_i \mid T)$ is the prior probability, representing how likely the target object $T$ would appear near object $O_i$, based on their semantic associations and known co-occurrence patterns. 
The term $P(D \mid O_i, T)$ is the likelihood, quantifying the probability of observing the data $D$ under the assumption that the target $T$ is indeed spatially associated with $O_i$. 
Additionally, we denote with $B_{none}$ hypothesis  that the target object is not present within the space we are exploring.

Given this theoretical representation, LLM is used to estimate the probability $P(O_i| D,T)$ for each detected object. Initially, $D$ contains only the user specification $U$ and the set of initially detected objects $\mathcal{O}$, and is expanded as the search progresses. The construction of a suitable LLM textual prompt is described in the following subsection.
Once the probabilities are estimated for each candidate $O_i$ using LLM, as the most relevant we pick the object that maximizes this posterior:
\begin{equation}
    O^* = \arg\max_{O_i \in \mathcal{O}} P(O_i \mid D, T)
\end{equation}
Given $O^*$, the decision logic that drives the UAV behavior is described as follows and given in Fig.~\ref{fig:overal_shema}. 
If the probability for the most relevant object $O^*$ is lower than a predefined minimal threshold, the UAV defaults to a predefined scanning strategy, such as wall following, with the goal of gathering more data. If the probability for the most relevant object $O^*$ exceeds the minimum threshold, the UAV initiates a thorough local search around the object. During the thorough search, the UAV continues to detect new objects and estimate their probabilities. If the estimated probability of a detected object increases above a certain high threshold, this means that the target object $T$ has been found, concluding the search process.
If the UAV does not find the target $T$ after doing a thorough search around the object $O_i$, the value of the total knowledge $D$ is updated and will lead to a significant decrease in the probability $P(O_i \mid T,D)$, which results in a redistribution of the probabilities to other remaining objects.

The overall system pipeline, from the input image $I_k$, occupied space $S_{occ}$, target object $T$ and user instructions $U$, to the selection of the most relevant object $O^*$, can be formally expressed as:
\begin{equation} 
PR( PE(VP(I_k), S_{occ}),T,D) \rightarrow O^*
\end{equation}

\subsubsection{Text prompt}
To effectively leverage the semantic reasoning capabilities of the LLM, it is essential to construct a prompt that provides both contextual object information and a clear interpretation of the user's search intent. The constructed prompt comprises two main components: (1) system and user information, which specifies the required reasoning task for the LLM and ensures it understands the context of the mission, and (2) the user’s target description combined with the observed environmental data, including detected objects and their spatial positions, which provides the LLM with the necessary situational details to reason effectively.

\begin{llmpromptbox}
\textbf{[System information]} "You're a helpful assistant that uses provided descriptions and observed objects to logically infer the probable location of a target object." 

\vspace{0.5em}
\textbf{[User information]:} "Given the target object description and the list of detected objects with their 3D coordinates, return the object most likely located near the target object."

\vspace{0.5em}
\textbf{[Input data]} \\
\texttt{User input}: Natural language user description of the target object to be located.\\
\texttt{Objects}: Objects detected within the environment along with their corresponding 3D coordinates.
\texttt{Visited}: Region of $S$ surrounding previously selected relevant objects that has been thoroughly scanned during the search process.

\vspace{0.5em}
\textbf{[Response format]:} \\
\texttt{Flag}: Indicates whether any of the detected objects are relevant to the current search query (exceeding minimal threshold).\\
\texttt{Most relevant}: The object with the highest likelihood of having the target object in its vicinity, specified by its label and 3D coordinates ($O^*$).\\
\texttt{Explanation}: The LLM's reasoning for selecting the identified object as the most relevant.
\end{llmpromptbox}

\subsection{Trajectory planning and thorough search}
\begin{figure}[]
    \centering
    \includegraphics[width=\linewidth]{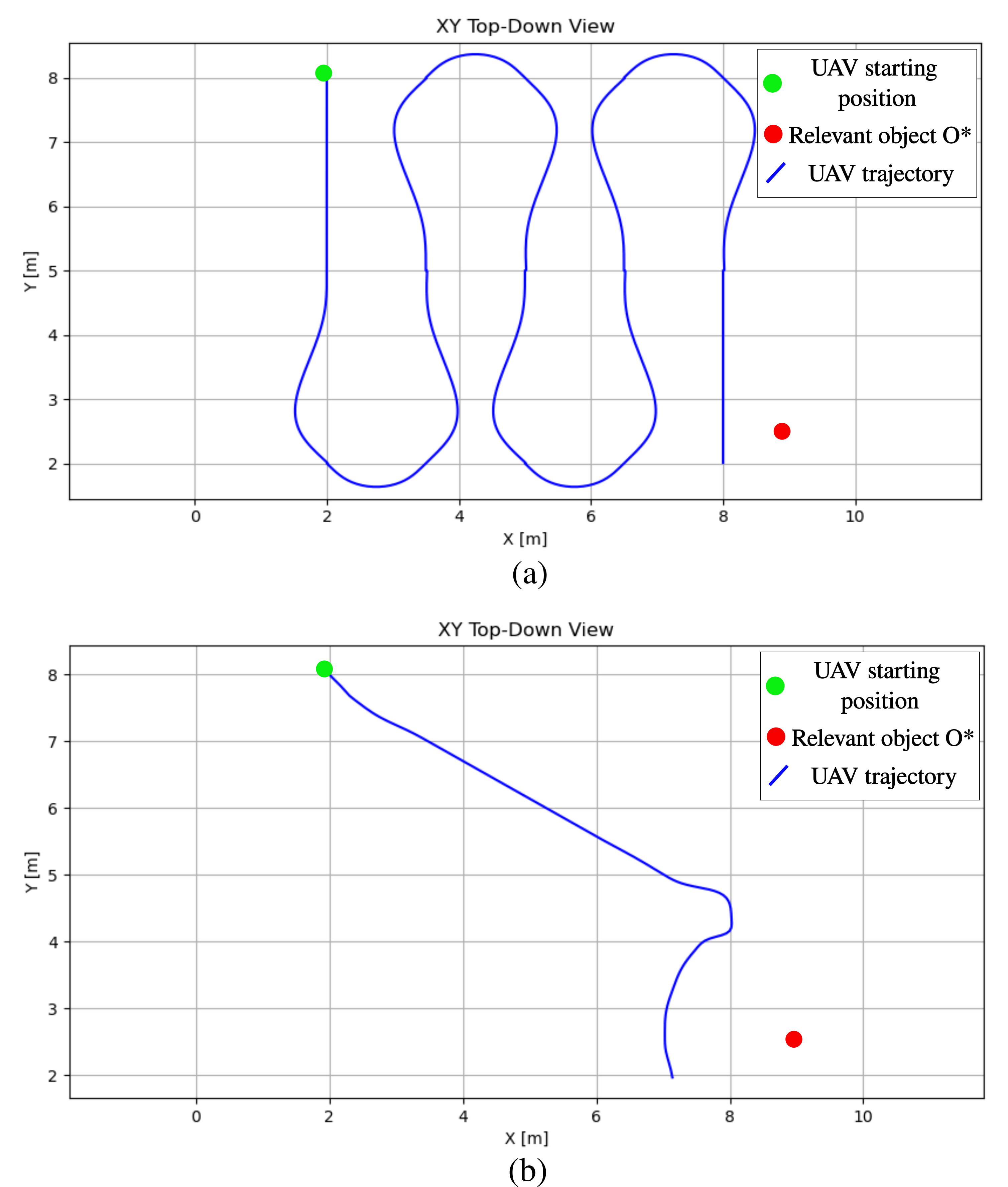}
    \caption{Top-down 2D visualization of generated trajectories: (a) Baseline Lawn Mower scanning trajectory; (b) Thorough search extending from the UAV’s current position $(1.9, 8.1, 2.0)$ to, and around, the relevant object located at $(9.0, 2.5, 1.1)$. Since the object is positioned near walls along $x=10$ and $y=0$, the circular inspection path is adapted to a semicircular trajectory, maintaining a safe distance from obstacles to ensure the UAV remains within the free navigable space.}
    \label{fig:thorough search}
\end{figure}

After the LLM identifies the most relevant object based on the current scene and user instruction, the UAV performs a thorough local search in its surrounding area, under the assumption that the target object may be located nearby. 
Since the object of interest could be partially or fully occluded, it is crucial to observe the region from multiple viewpoints. 
To achieve this, a circular inspection trajectory is generated around the selected object at a specified radius and fixed altitude, enabling the UAV to scan the area from all angles and improve detection reliability. 
As illustrated in Fig.~\ref{fig:thorough search}(b), the trajectory is generated from the UAV’s current position to, and around, the relevant object, while omitting any waypoints that would intersect walls or lie outside the free space. 
The performance of the proposed method is evaluated against a baseline lawn-mower search strategy, which systematically covers the entire area using a uniform back-and-forth sweeping pattern, as shown in Fig.~\ref{fig:thorough search}(a).

To navigate efficiently through the environment while avoiding obstacles, the UAV employs a two-stage trajectory planning approach.
(1) Discrete path planning using A* search\cite{Hart1968Astar} in 3D space to reach the region of interest. Prior to executing a circular scanning maneuver, the system queries the OctoMap to ensure that the planned trajectory lies within free space and does not intersect with any obstacles.
(2) Continuous trajectory generation over 3D waypoints is then achieved using 7th-order polynomial splines, enabling smooth, dynamically feasible flight paths that satisfy constraints on position, velocity, and acceleration.

During the thorough search around the selected object $O^*$, the UAV continuously gathers new visual data and maintains a record of visited areas. This information is used to update the probabilistic relevance module, which iteratively reassesses the scene and determines the next best action if the target object has not yet been found.

\section{Results}
\label{sec:results}
\begin{figure*}[t!]
    \centering
    \includegraphics[width=\linewidth]{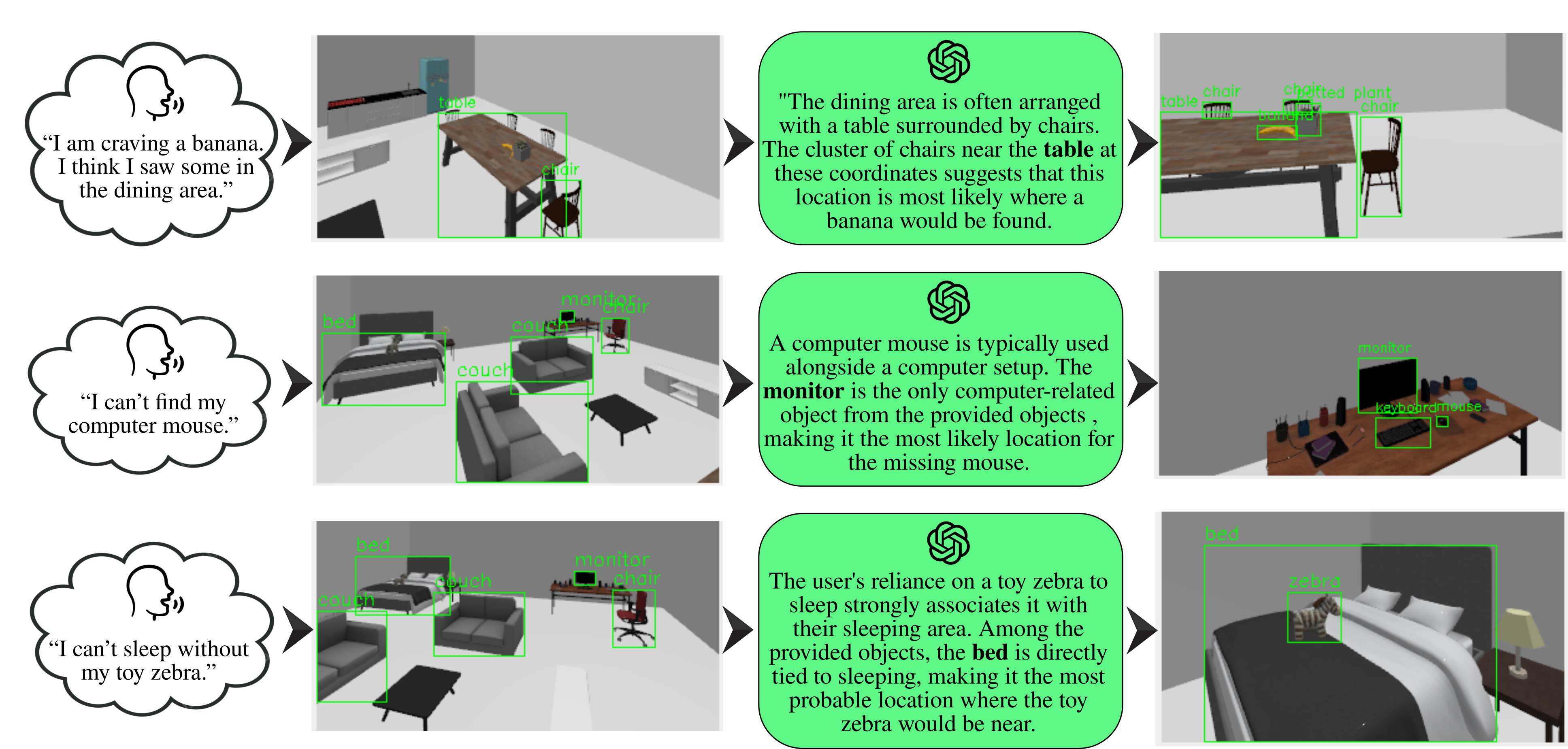}
    \caption{Illustrated simulation test scenarios: the process begins with the user providing natural-language instructions, after which the UAV performs a 360-degree scan to survey and detect surrounding objects. The large language model (LLM) then reasons over the gathered information, determines the next action, and generates an explanation for the bolded relevant object. Finally, the UAV executes a thorough-search routine around the relevant object until the specified target object is successfully located.}
    \label{fig:scenarios}
\end{figure*}

The results of this work are presented through a detailed analysis of both simulation and real-world experiments, highlighting the effectiveness of semantic-guided search in comparison to traditional, non-semantic search strategies.
The experimental platform is the Crazyflie 2.1 nano-UAV, a lightweight and modular quadcopter equipped with an AI Deck that features a monochrome camera. 
Due to its limited payload capacity, the Crazyflie cannot carry advanced sensors such as LiDAR for real-time mapping. 
Therefore, all environments, both simulated and real, were pre-mapped using Cartographer SLAM with point clouds acquired from an external LiDAR sensor. 
This pre-built map was then used for navigation and semantic grounding. 
Since both Cartographer SLAM and OctoMap support online mapping, the presented system can readily extend to larger UAV platforms equipped with onboard LiDAR sensors, enabling incremental mapping and exploration of unknown environments in real-time.
During flight, the UAV’s camera images were processed offboard using YOLO11\cite{yolo11_ultralytics} for object detection, and detected objects were grounded in the map using OctoMap's ray-casting techniques. 
The system’s ability to interpret instructions and object context was powered by OpenAI's o3-mini large language model, allowing semantic reasoning about likely object locations.

\subsection{Simulation results}
\begin{table*}[htbp]
\centering
\caption{Performance Comparison Between LLM-Guided Search and Lawn Mower Baseline Across Three Scenarios}
\label{tab:results}
\begin{tabular}{@{}lccccc@{}}
\toprule
\textbf{Scenario} & \textbf{Method} & \textbf{Success Rate (\%)} & \textbf{Search Time Improvement (\%)} & \textbf{Path Length Reduction (\%)} \\
\midrule
\multirow{2}{*}{Banana} 
    & LLM-guided & 100 & \multirow{2}{*}{36.02 ± 30.52} & \multirow{2}{*}{52.26 ± 38.16} \\
    & Lawn Mower & 65  &                                   &                              \\
\midrule
\multirow{2}{*}{Computer Mouse}
    & LLM-guided & 100 & \multirow{2}{*}{24.83 ± 25.13} & \multirow{2}{*}{43.50 ±  33.79} \\
    & Lawn Mower & 75  &                                   &                              \\
\midrule
\multirow{2}{*}{Toy Zebra}
    & LLM-guided & 100 & \multirow{2}{*}{29.24 ± 28.71} & \multirow{2}{*}{51.93 ±  27.86} \\
    & Lawn Mower & 95  &                                   &                              \\
\bottomrule
\end{tabular}
\vspace{0.2cm}
\end{table*}

We evaluated the proposed LLM-guided semantic search framework against a baseline lawn mower scanning strategy across, as seen in Fig.\ref{fig:scenarios}, three object-search scenarios:
\begin{itemize}
    \item Scenario I: Banana - partially occluded by a vase, requiring inference about likely locations (e.g. dining area)
    \item Scenario II: Computer mouse - requiring inference about likely locations based on co-occurrence of objects (e.g. part of office setup). 
    \item Scenario III: Toy zebra - with and without detailed user instructions to test the impact of semantic context.
\end{itemize}
Each scenario was tested from 20 randomly selected UAV starting positions in a simulated $10 \times 10$ m environment using a Gazebo Garden simulator, totaling 120 flights (60 LLM-guided, 60 lawn mower). Performance was assessed using three metrics:
\begin{itemize}
    \item Success Rate (SR) - The proportion of search trials in which the target object was successfully localized by the UAV, expressed as a percentage.
    \item Search Time Improvement (STI) - The percentage reduction in elapsed time from takeoff to successful object localization, relative to the baseline method. 
    \item Path Length Reduction (PLR) - The percentage reduction in the total distance traveled by the UAV from takeoff to object localization, relative to the baseline method.
\end{itemize}
The measured duration of each search accounts for the total duration required to execute all system modules and is not directly proportional to the distance traversed by the UAV. 
Following liftoff, the UAV performs a 360-degree scan of the environment to obtain an initial overview of the scene. 
In the proposed method, the reported search time also includes the time required by the LLM to reason and determine the first action following this initial scan, with an average reasoning time of approximately 9.97 seconds.
In subsequent decisions, the reasoning process is executed concurrently with the UAV’s trajectory execution and therefore does not add to the overall search time.

Table \ref{tab:results} summarizes the aggregate performance metrics of the LLM-guided semantic search framework across all randomized starting positions, benchmarked against a conventional lawn mower baseline. 
For each scenario, the table reports the Success Rate (SR) alongside statistically significant improvements in search time and path length, expressed as mean percentage reductions with standard deviations $\sigma$. 
These improvements were computed exclusively over trials where both methods successfully localized the target object, ensuring fair comparison. 

\begin{figure*}[t!]
    \centering
    \includegraphics[width=\linewidth]{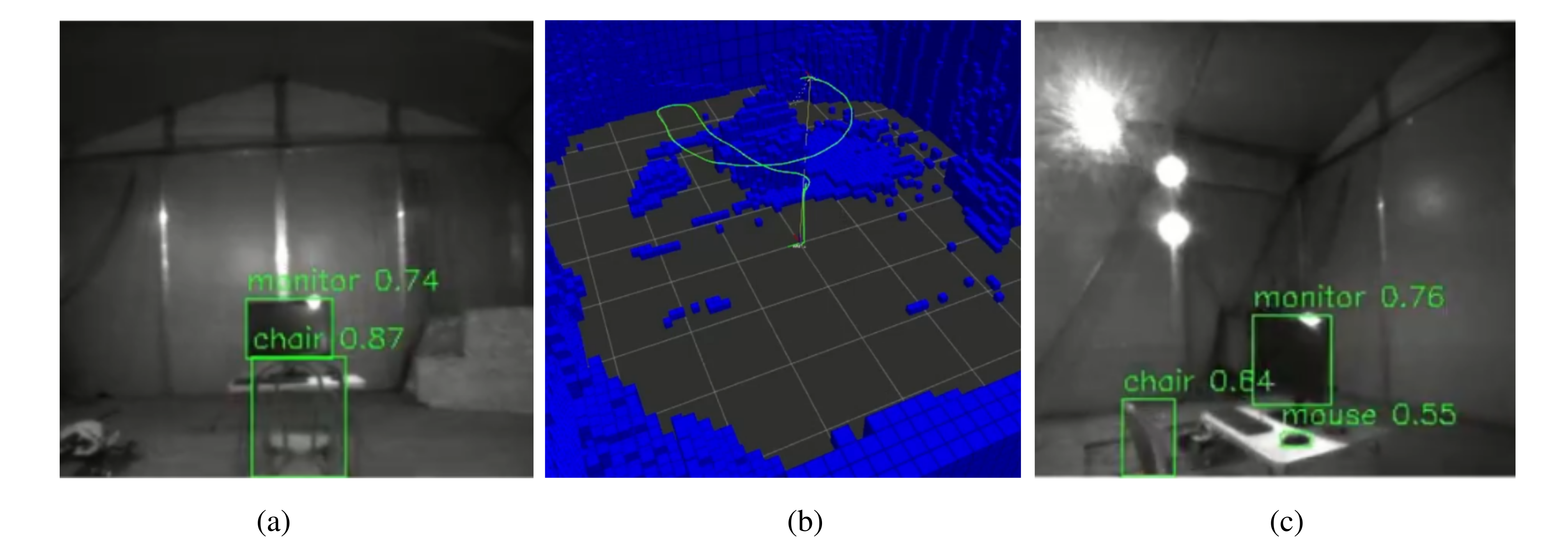}
    \caption{Real-world experiment with the target object set as a computer mouse: (a) Initial camera view of the scene, where the target object is occluded but the computer monitor is visible; (b) OctoMap representation, with occupied space visualized by blue voxels and the UAV’s executed trajectory shown in green; (c) Camera view during the thorough search process, where the computer mouse becomes detectable, leading to successful completion of the search.}
    \label{fig:experiment}
\end{figure*}
\subsubsection{Scenario I - Banana}
Given the following user instruction $U$: “I am craving a banana. I think I saw some in the dining area”, the LLM-guided approach identified banana as a search object $T$, and achieved a 100\% success rate, compared to 65\% for the lawn mower baseline. 
The LLM-guided UAV reliably guided the UAV to perform a thorough search around the dining table, which was selected as having the highest probability of finding a banana ($O^*$), and localized the target despite frequent occlusions. Other detected objects were office table, chairs, bed, and others. 
On average, the LLM-guided method reduced the search time by 36.02\% ($\sigma=30.52$) and the path length by 52.26\% ($\sigma=38.16$) relative to the baseline.

\subsubsection{Scenario II - Computer Mouse}
In this scenario, the instruction $U$ was “I can’t find my computer mouse”. The LLM-guided UAV achieved a 100\% success rate, outperforming the baseline’s 75\%. 
By leveraging semantic priors, the system was able to infer and prioritize likely search locations such as desks, monitors, and other office-related areas, thus focusing the search on regions with the highest probability of success. 
This targeted reasoning enabled the UAV to avoid unnecessary exploration of irrelevant spaces, resulting in a 24.83\% reduction in search time ($\sigma=25.13$) and a 43.50\% reduction in path length ($\sigma=33.79$) compared to the lawn mower approach.

\subsubsection{Scenario III - Toy Zebra}
With the prompt, “I can’t sleep without my toy zebra”, both methods achieved high performance; however, the LLM-guided search maintained a perfect 100\% success rate, slightly exceeding the baseline’s 95\%. 
Notably, the effectiveness of the LLM-guided approach in this scenario was directly attributable to the user’s detailed description specifying that the zebra was a toy associated with sleep, enabling the model to infer likely search regions such as bedrooms or resting areas. 
Without such contextual information, the LLM would lack the semantic grounding necessary to prioritize relevant locations, as a zebra alone has no inherent connection to an indoor environment. 
This result underscores the critical role of user-provided context in guiding efficient search. 
Even in this otherwise ambiguous case, the LLM-guided method reduced search time on average by 29.24\% ($\sigma=28.71$) and path length by 51.93\% ($\sigma=27.86$) compared to the baseline.

The results demonstrate the framework’s robustness in reducing mission duration and trajectory complexity while maintaining perfect search accuracy, even in occluded or context-dependent scenarios. 

\subsection{Experimental results}
Following successful simulation results, the developed semantic search algorithm was validated on a real Crazyflie 2.1 nano-UAV equipped with an AI Deck featuring a monochrome camera. 
For robust tracking within the flight arena (8 $\times$ 10 meters), a reflective marker was mounted atop the UAV and the OptiTrack motion capture system was used for precise localization. 
The environment consisted of a table and chair, with a monitor, keyboard, and computer mouse placed on the table. 
Prior to flight, the space was mapped using LiDAR and Cartographer SLAM to generate an OctoMap for navigation and object grounding. 
Communication with the UAV was established via CrazyRadio for control and WiFi for image streaming. 
The UAV received the instruction “Where is my computer mouse?”, with the LLM clarifying the search target as a computer accessory. 
Upon takeoff, the UAV detected the table, chair, and monitor, and, guided by semantic reasoning, focused its search around the monitor, successfully localizing the mouse. 

This experiment demonstrates that the proposed LLM-guided framework is feasible and effective for real-time semantic search on a lightweight UAV in a real environment.
\section{Conclusion}

This paper presents a novel framework that integrates large language models into the control loop of UAVs, transforming object-search missions from exhaustive, time-consuming sweeps into targeted, context-aware explorations. 
By continuously reasoning over real-time detections, spatial maps, and natural-language instructions, the proposed system achieves 100\% mission success across diverse scenarios, while reducing average mission duration by up to 36\% and path length by over 50\% compared to conventional lawn-mower search patterns. 
The incorporation of a thorough-search routine further ensures robust and reliable localization of targets, regardless of their size or occlusion, delivering consistent performance across varying object classes and positions.
The framework leverages lightweight object detectors in combination with an LLM to enable real-time semantic inference and perception, avoiding the computational overhead associated with vision language models. 
The system was successfully validated on a UAV in a real-world setting, demonstrating its practical effectiveness, robustness, and adaptability under realistic operational conditions.

Future work will focus on refining prompt engineering strategies and semantic priors, as well as integrating open-set object detectors to accelerate inference and support zero-shot generalization in unstructured and previously unseen environments.

\bibliographystyle{ieeetr}
\bibliography{Bibliography/bibliography}

@inproceedings{anderson2018,
  title={Vision-and-language navigation: Interpreting visually-grounded navigation instructions in real environments},
  author={Anderson, Peter and Wu, Qi and Teney, Damien and Bruce, Jake and Johnson, Mark and S{\"u}nderhauf, Niko and Reid, Ian and Gould, Stephen and Van Den Hengel, Anton},
  booktitle={Proceedings of the IEEE Conference on Computer Vision and Pattern Recognition (CVPR)},
  pages={3674--3683},
  year={2018}
}

@misc{anderson2018objgoal,
      title={On Evaluation of Embodied Navigation Agents}, 
      author={Peter Anderson and Angel Chang and Devendra Singh Chaplot and Alexey Dosovitskiy and Saurabh Gupta and Vladlen Koltun and Jana Kosecka and Jitendra Malik and Roozbeh Mottaghi and Manolis Savva and Amir R. Zamir},
      year={2018},
      eprint={1807.06757},
      archivePrefix={arXiv},
      primaryClass={cs.AI},
      url={https://arxiv.org/abs/1807.06757}, 
}

@inproceedings{NEURIPS2020,
 author = {Chaplot, Devendra Singh and Gandhi, Dhiraj Prakashchand and Gupta, Abhinav and Salakhutdinov, Russ R},
 booktitle = {Advances in Neural Information Processing Systems},
 editor = {H. Larochelle and M. Ranzato and R. Hadsell and M.F. Balcan and H. Lin},
 pages = {4247--4258},
 publisher = {Curran Associates, Inc.},
 title = {Object Goal Navigation using Goal-Oriented Semantic Exploration},
 url = {https://proceedings.neurips.cc/paper_files/paper/2020/file/2c75cf2681788adaca63aa95ae028b22-Paper.pdf},
 volume = {33},
 year = {2020}
}

@inproceedings{habitat,
  title={Habitat: A platform for embodied ai research},
  author={Savva, Manolis and Kadian, Abhishek and Maksymets, Oleksandr and Zhao, Yili and Wijmans, Erik and Jain, Bhavana and Straub, Julian and Liu, Jia and Koltun, Vladlen and Malik, Jitendra and others},
  booktitle={Proceedings of the IEEE/CVF international conference on computer vision},
  pages={9339--9347},
  year={2019}
}

@inproceedings{gibson,
  title={Gibson env: Real-world perception for embodied agents},
  author={Xia, Fei and Zamir, Amir R and He, Zhiyang and Sax, Alexander and Malik, Jitendra and Savarese, Silvio},
  booktitle={Proceedings of the IEEE conference on computer vision and pattern recognition},
  pages={9068--9079},
  year={2018}
}

@article{matterport3d,
  title={Matterport3d: Learning from rgb-d data in indoor environments. arXiv 2017},
  author={Chang, A and Dai, A and Funkhouser, T and Halber, M and Niessner, M and Savva, M and Song, S and Zeng, A and Zhang, Y},
  journal={arXiv preprint arXiv:1709.06158}
}

@article{majumdar2023zson,
  title={Zson: Zero-shot object-goal navigation using multimodal goal embeddings},
  author={Majumdar, Arjun and Aggarwal, Gunjan and Devnani, Bhavika and Hoffman, Judy and Batra, Dhruv},
  journal={Advances in Neural Information Processing Systems},
  volume={35},
  pages={32340--32352},
  year={2022}
}

@inproceedings{gadre2022lzson,
  title={Cows on pasture: Baselines and benchmarks for language-driven zero-shot object navigation},
  author={Gadre, Samir Yitzhak and Wortsman, Mitchell and Ilharco, Gabriel and Schmidt, Ludwig and Song, Shuran},
  booktitle={Proceedings of the IEEE/CVF Conference on Computer Vision and Pattern Recognition},
  pages={23171--23181},
  year={2023}
}

@inproceedings{liu2023aerialvln,
  title={Aerialvln: Vision-and-language navigation for uavs},
  author={Liu, Shubo and Zhang, Hongsheng and Qi, Yuankai and Wang, Peng and Zhang, Yanning and Wu, Qi},
  booktitle={Proceedings of the IEEE/CVF International Conference on Computer Vision},
  pages={15384--15394},
  year={2023}
}

@inproceedings{
kuang2024openfmnav,
title={Open{FMN}av: Towards Open-Set Zero-Shot Object Navigation via Vision-Language Foundation Models},
author={Yuxuan Kuang and Hai Lin and Meng Jiang},
booktitle={ICLR 2024 Workshop on Large Language Model (LLM) Agents},
year={2024},
url={https://openreview.net/forum?id=l1eI6NrXYX}
}

@inproceedings{
shah2022lmnav,
 title={LM-Nav: Robotic Navigation with Large Pre-Trained Models of Language, Vision, and Action}, 
      author={Dhruv Shah and Blazej Osinski and Brian Ichter and Sergey Levine},
booktitle={Proceedings of The 6th Conference on Robot Learning, PMLR 205:492-504},
year={2022},
url={https://proceedings.mlr.press/v205/shah23b/shah23b.pdf}
}

@inproceedings{cai2023pixnav,
  title={Bridging zero-shot object navigation and foundation models through pixel-guided navigation skill},
  author={Cai, Wenzhe and Huang, Siyuan and Cheng, Guangran and Long, Yuxing and Gao, Peng and Sun, Changyin and Dong, Hao},
  booktitle={2024 IEEE International Conference on Robotics and Automation (ICRA)},
  pages={5228--5234},
  year={2024},
  organization={IEEE}
}

@ARTICLE{objectnaw_review,
  author={Sun, Jingwen and Wu, Jing and Ji, Ze and Lai, Yu-Kun},
  journal={IEEE Transactions on Automation Science and Engineering}, 
  title={A Survey of Object Goal Navigation}, 
  year={2025},
  volume={22},
  number={},
  pages={2292-2308},
  doi={10.1109/TASE.2024.3378010}}

@inproceedings{zhou2023esc,
  title={Esc: Exploration with soft commonsense constraints for zero-shot object navigation},
  author={Zhou, Kaiwen and Zheng, Kaizhi and Pryor, Connor and Shen, Yilin and Jin, Hongxia and Getoor, Lise and Wang, Xin Eric},
  booktitle={International Conference on Machine Learning},
  pages={42829--42842},
  year={2023},
  organization={PMLR}
}

@misc{crazyflie21,
  author       = {Bitcraze},
  title        = {Crazyflie 2.1 Product Page},
  note         = {Accessed: 10. studenog 2024.},
  url          = {https://www.bitcraze.io/products/old-products/crazyflie-2-1/}
}

@software{yolo11_ultralytics,
  author = {Glenn Jocher and Jing Qiu},
  title = {Ultralytics YOLO11},
  version = {11.0.0},
  year = {2024},
  url = {https://github.com/ultralytics/ultralytics},
  orcid = {0000-0001-5950-6979, 0000-0002-7603-6750, 0000-0003-3783-7069},
  license = {AGPL-3.0}
}

@inproceedings{ester1996dbscan,
  title={A density-based algorithm for discovering clusters in large spatial databases with noise.},
  author={Ester, Martin and Kriegel, Hans-Peter and Sander, Jorg and Xu, Xiaowei and others},
  booktitle={International Conference on Knowledge Discovery and Data Mining (KDD)},
  volume={96},
  number={34},
  pages={226--231},
  year={1996}
}

@article{Hart1968Astar,
  doi = {10.1109/tssc.1968.300136},
  url = {https://doi.org/10.1109/tssc.1968.300136},
  year = {1968},
  publisher = {Institute of Electrical and Electronics Engineers ({IEEE})},
  volume = {4},
  number = {2},
  pages = {100--107},
  author = {Peter Hart and Nils Nilsson and Bertram Raphael},
  title = {A Formal Basis for the Heuristic Determination of Minimum Cost Paths},
  journal = {{IEEE} Transactions on Systems Science and Cybernetics}
}

\end{document}